\documentclass[a4paper,twoside,anonymous=True]{article}

\usepackage{epsfig}
\usepackage{subcaption}
\usepackage{calc}
\usepackage{amssymb}
\usepackage{amstext}
\usepackage{amsmath}
\usepackage{amsthm}
\usepackage{multicol}
\usepackage{pslatex}
\usepackage{apalike}
\usepackage{breqn}
\usepackage{booktabs}
\usepackage{array}
\usepackage{tabularx}
\usepackage{SCITEPRESS}     % Please add other packages that you may need BEFORE the SCITEPRESS.sty package.

\begin{document}

\title{How does AI play football? \\An analysis of RL and real-world football strategies}

% \author{Atom Scott\inst{1}\orcidID{0000-1111-2222-3333} \and
% Masaki Onishi\inst{1}\orcidID{2222--3333-4444-5555}}
% %
% \authorrunning{A. Scott et al.}
% \institute{National Institute of Advanced Industrial Science and Technology (AIST) \\
% \email{atom.scott@aist.go.jp}
% \email{onishi@ni.aist.go.jp}
% }
%

% \author{
%     \authorname{Anonymous}
% }
% \author{
%     \authorname{
%     Atom Scott\sup{1}\orcidAuthor{0000-0003-2500-1096}, 
%     Keisuke Fujii\sup{2}\orcidAuthor{0000-0002-4580-4868} and 
%     Masaki Onishi\sup{1}\orcidAuthor{0000-0001-5487-4297}}
    
% \affiliation{\sup{1}National Institute of Advanced Industrial Science and Technology (AIST), Tsukuba, Japan}
% \affiliation{\sup{2}Graduate School of Informatics, Nagoya University, Nagoya, Japan}
% \email{atom.scott@aist.go.jp, fujii@i.nagoya-u.ac.jp, onishi@ni.aist.go.jp}
% }
\author{
    \authorname{
    Atom Scott\sup{1}, 
    Keisuke Fujii\sup{2} and 
    Masaki Onishi\sup{1}}
    
\affiliation{\sup{1}National Institute of Advanced Industrial Science and Technology (AIST), Tsukuba, Japan}
\affiliation{\sup{2}Graduate School of Informatics, Nagoya University, Nagoya, Japan}
\email{atom.scott@aist.go.jp, fujii@i.nagoya-u.ac.jp, onishi@ni.aist.go.jp}
}


\abstract{Recent advances in reinforcement learning (RL) have made it possible to develop sophisticated agents that excel in a wide range of applications. Simulations using such agents can provide valuable information in scenarios that are difficult to scientifically experiment in the real world. In this paper, we examine the play-style characteristics of football RL agents and uncover how strategies may develop during training. The learnt strategies are then compared with those of real football players. We explore what can be learnt from the use of simulated environments by using aggregated statistics and social network analysis (SNA). As a result,  we found that (1) there are strong correlations between the competitiveness of an agent and various SNA metrics and (2) aspects of the RL agents play style become similar to real world footballers as the agent becomes more competitive.  We discuss further advances that may be necessary to improve our understanding necessary to fully utilise RL for the analysis of football.}

% However, there has been little attempt to justify  and compare the different characteristics between learnt agents of increasing competitiveness, and expert human performers. 

\onecolumn \maketitle \normalsize \setcounter{footnote}{0} \vfill

\section{\uppercase{Introduction}}
\label{sec:introduction}

Over the last decade there has been an increase in interest towards analytics in football (soccer), and many other team-sports. Increasing compute power and data has added to the effectiveness of statistical analysis and more importantly, allowed for compute-intensive and data-intensive machine learning methods. Many success stories have been well documented in mainstream publications such as “The Numbers Game” \cite{Anderson2013TheWrong}, “Basketball on Paper” \cite{Oliver2020BasketballAnalysis} and perhaps most well known, “Moneyball” \cite{Lewis2004Moneyball:Game}. As a result, a growing number of sports teams now adopt specialist roles for analytics. If we assume such trends are to continue, it is likely both compute power and the amount of available data will exponentially increase in forthcoming years. However, it will remain nearly impossible to collect real-world sport data in a scientific manner where variables can be controlled. This can not be helped since top level sports are highly competitive in nature and leave very little room for experimentation. To solve this problem, agent-based simulation (ABS) can be used as a test-bed to simulate various scenarios in a scientific manner. 

\begin{figure}[!tbp]
  \centering
  \begin{minipage}[b]{0.45\textwidth}
    \includegraphics[width=\textwidth]{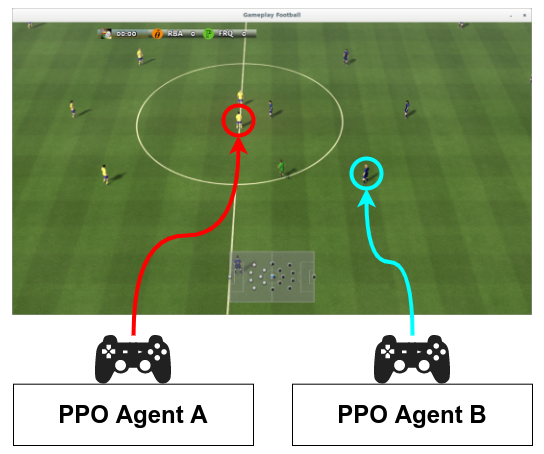}
    \caption{A representation of the agent setup where a single RL agent is used to control a single active player of a team. The illustration shows an image of the rendered environment \cite{Kurach2019GoogleEnvironment} with arrows pointing to the active-players. Active players can be switched in-game to and from non-active players that are controlled via another in-game rule based system.}\label{rep}
  \end{minipage}
\end{figure}

Recently, deep reinforcement learning (RL) methods have shown it is possible to train agents, from scratch, that outperform human experts in both traditional \cite{Silver2016MasteringSearch,silver2017mastering} and modern games \cite{Mnih2013PlayingLearning,alphastarblog,Berner2021DotaLearning}. These breakthroughs, coupled with increasingly sophisticated simulation environments, are a promising new direction of analysis in sports. Therefore in this paper, we examine the characteristics of football playing RL agents and uncover how strategies may develop during training. Out of the many team sports that exist we choose to focus on football due to its popularity and the availability of a sufficient simulation environment (see \S2 for more detail). We use the Google Research Football environment  \cite{Kurach2019GoogleEnvironment} to train football playing RL agents in a single agent manner. Fig.~\ref{rep} illustrates a representation of the training setup we used. Another problem concerning the use of ABS is that the domain gap between RL-agents and real-world football players is not clear. To gain a better understanding of this domain gap,  we compared the characteristics of football strategies in RL agents and real-world football players. In summary, the main contributions of the study are as follows: 

\begin{itemize}
    \item We compared the characteristics of football playing RL agents \cite{Kurach2019GoogleEnvironment} in various training processes and real-world football players for the first time, thus verifying simulations as a practical approach for football analysis.
    
    \item We found that more competitive RL agents have a more similar and well-balanced passing strategy to real-world footballers in comparison to less competitive RL agents. 

    \item We analyzed how the football strategies of RL-agents evolve as the  competitiveness of the agent increases. Strong correlations were found between many aggregated statistics / social network analysis and the competitiveness of the agent. 
    
    % which contribute to the understanding of RL agents from a more complex football perspective. \\
    
    % \item We found that RL agents play more similarly to real world football players in correlation to the competitiveness of the agent, implying that better learning strategies are crucial to the practical use of RL agents in sports analytics.\\
\end{itemize}

\noindent The outline of this paper is as follows. \S2 provides background on agent-based simulation, deep RL and football analytics. \S3 and \S4 discuss the preliminaries and methods used to train deep RL-agents and the metrics used to analyse playing characteristics. We present results and discussions in \S5. Finally, we summarise our conclusions and future work in \S6.

\section{\uppercase{Related Works}}
% The following three subsections provide an overview of each area that is essential to our work.

\subsection{Agent-Based Simulation}
Agent-based simulation (ABS) is a computationally demanding technique for simulating dynamic complex systems and observing “emergent” behaviour. With the use of ABS, we can explore different outcomes of phenomena where it is infeasible to conduct research testing and hypothesis formulations in real life. In the context of football we can use ABS to examine effects of different formations on match outcomes or study various play styles using millions of simulated football games. 
The availability of good simulation environments are critical to ABS. Fortunately, football has received a lot of attention in this field thanks to the long history of the RoboCup simulation track \cite{itsuki1995soccer}. In recent years, many other simulation environments have also been introduced \cite{Liu2019EmergentCompetition,Cao2020REINFORCEMENTCRITICS,Liu2021FromFootball}. Amongst others, the Google Research Football environment \cite{Kurach2019GoogleEnvironment} stands out as an interesting test-bed. Kaggle has held a competition with over a thousand teams participating\footnote{https://www.kaggle.com/c/google-football} and researchers have already started to develop methods to analyze football matches using Google Research Football via graphical tools \cite{PinciroliVago2020INTEGRA:Matches} or RL inspired metrics \cite{Garnier2021EvaluatingLearning}. 
Therefore we choose to use the Google Research Football environment to conduct our simulations. It reproduces a full football match with all of its usual regulations and events, as well as player tiredness, misses, etc. We list an overview of available simulation environments in Table~\ref{sim-table}).

\begin{table*}[]
\caption{An overview of various football simulation environments.}\label{sim-table}
\begin{tabular}{| p{0.25\linewidth} | p{0.65\linewidth} |}
    \hline
    Environment & Description \\  \hline
    RoboCup Soccer \cite{itsuki1995soccer} &
        An 11 vs 11 soccer simulator. Agents receive noisy input from virtual sensors and perform some basic commands such as dashing, turning or kicking.\\ \hline
    MuJoCo 2 vs 2 \cite{Liu2019EmergentCompetition} &   
        A 2 vs 2 football environment with simulated physics built on MuJoCo \cite{Todorov2012MuJoCo:Control}.
        Uses relatively simple bodies with a 3-dimensional action space.\\ \hline
        
    Unity 2 vs 2 \cite{Cao2020REINFORCEMENTCRITICS}&   
        A 2 vs 2 football environment built on unity. Two types of players with slightly different action spaces are available.\\ \hline
    
    Google Research \cite{Kurach2019GoogleEnvironment} &   
        An 11 vs 11 soccer environment built on GameplayFootball. Simulates a full football game and includes common aspects such as goals, fouls, corners, etc. \\ \hline

    Humanoid \cite{Liu2021FromFootball} &
        A 2 vs 2 football environment with simulated physics built on MuJoCo \cite{Todorov2012MuJoCo:Control} designed to embed sophisticated motor control of the humanoid. Physical aspects such as the radius of the ball and goal size are adjusted in proportion to the height of the humanoid.\\  \hline
\end{tabular}
\end{table*}

\subsection{Deep Reinforcement Learning}
Deep RL is a subset of RL that combines the traditional reinforcement learning setup, in which agents learn optimal actions in a given environment, with deep neural networks. There have been many remarkable examples of agents trained via deep RL outperforming experts. A remarkable example of this is Deepmind’s AlphaGo \cite{Silver2016MasteringSearch}. Its successors AlphaZero \cite{silver2018general} and Muzero \cite{Schrittwieser2020MasteringModel}  achieved a superhuman level of play in the games of chess, shogi and go solely via self-play.
 
In contrast to the single-player, deterministic, perfect information setup for the classical games mentioned above, football is a highly stochastic imperfect information game with multiple players that construct a team. Although these characteristics have made it difficult to learn through self-play, recent works have shown promising results in similar categorised games such as DotA and StarCraft. For example, OpenAI Five \cite{Berner2021DotaLearning} scaled existing RL systems to unprecedented levels, while performing “surgery” to utilise thousands of GPUs over multiple months. On the other hand, AlphaStar \cite{Vinyals2019GrandmasterLearning} populated a league consisting of agents with distinct objectives, and introduced agents that specifically try to exploit shortcomings in other agents and in the league. This allowed agents to train while continually adapting strategies and counter-strategies. 

As for research directly related to football, Robot soccer \cite{itsuki1995soccer} has been one of the longstanding challenges in AI. Although this challenge has been tackled with machine learning techniques \cite{Riedmiller2009ReinforcementSoccer,Macalpine2018Journal}, it has not yet been mastered by end-to-end deep RL. Nonetheless, baseline approaches for other simulation environments mostly utilise deep RL. \cite{Liu2019EmergentCompetition} used a 
population-based training with evolution and reward shaping on a recurrent policy with recurrent action-value estimator in MuJoCo Soccer. Whereas \cite{Cao2020REINFORCEMENTCRITICS} showed that RL from hierarchical critics was affected in the Unity 2 vs 2 environment. Proximal Policy Optimization (PPO) \cite{Schulman2017ProximalAlgorithms}, IMPALA \cite{Espeholt2018IMPALA:Architectures} and Ape-X DQN \cite{Horgan2018DistributedReplay} were provided as benchmark results for Google Research Football \cite{Kurach2019GoogleEnvironment}. Finally a combination of imitation learning, single and multi-agent RL and population-based training was used in Humanoid Football \cite{Liu2021FromFootball}.

Many researchers have attempted to model the behaviour of players by predicting the short term future contrary to the long-term horizon goal approach using deep RL \cite{le2017coordinated,Felsen2018WhereAutoencoders,Yeh_2019_CVPR}. Such research offers important insights into what architectures/time horizons/rewards may be effective. 

\subsection{Football Analytics}
Football has been considered to be one of the most challenging sports to analyze due to the number of players, continuous events and low frequency of points (goals). Therefore, it is only recently that a data-driven approach has started to gain attention. Nevertheless, numerous approaches, from the simple aggregation of individual/team play statistics \cite{Novatchkov2013ArtificialTraining}, to complex methods, such as those that use gradient boosting to model the value of actions \cite{decroos2018actions}. In general one can observe two different types of analysis. The first focuses on evaluating the overall performance of a single player or team. In this case, an action is usually valued then aggregated by either player or team. \cite{decroos2018actions} assigned values to on-ball action actions by measuring their effect on the probabilities that a team will score. In turn, \cite{fernandez2018wide} proposed a method to value off the ball actions by estimating pitch value with a neural network. The second category of analysis is strategy or play style analysis. Methods such as automatic formation \cite{Bialkowski2016DiscoveringData} or tactic \cite{Gyarmati2015AutomaticTeams,Decroos2018AutomaticData} discovery fall into this category. Social network analysis is also a well used method to analyse interactions between players \cite{Clemente2016SocialAnalysis,Buldu2018UsingGame}. Network metrics such as betweenness, centrality and eccentricity are often used. \cite{Pena2012AStrategies} demonstrated that winning teams presented lower betweenness scores. Similarly, \cite{Goncalves2017ExploringFootball} provided evidence that a lower passing dependency for a given player and higher intra-team well-connected passing relations may optimise team performance.

\section{\uppercase{Preliminaries}}

\begin{figure*}[t]
\includegraphics[width=\linewidth]{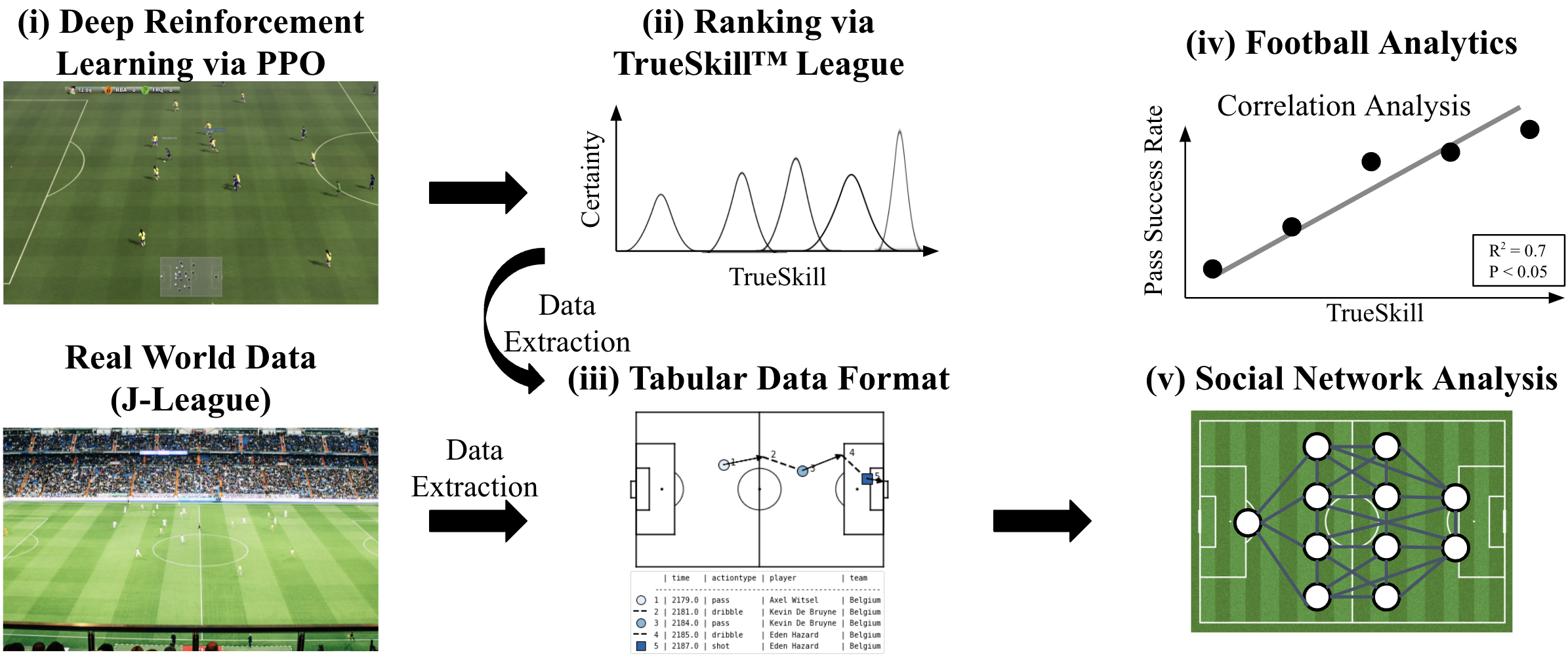}
\caption{An overview of the proposed framework. Details for steps (i) - (iv) are detailed in \S\ref{Agent Training and Ranking}, \S\ref{TrueSkill Ranking Implementation}, \S\ref{Data Extraction}, and \S\ref{Data Analysis} respectively. In (iii), data is converted to a tabular format inspired by SPADL \cite{Decroos2019ActionsSoccer}.} \label{overview} 
\end{figure*}

\subsection{Proximal Policy Optimization}
To learn policies for agents to play Google Research Football, we follow the original paper \cite{Kurach2019GoogleEnvironment} and use Proximal Policy Optimisation (PPO) \cite{Schulman2017ProximalAlgorithms}. PPO belongs to a family of reinforcement learning called policy gradient methods. These methods try to find an optimal behaviour strategy by alternating between optimising a clipped surrogate objective function and sampling data through interactions with the environment. The objective function of PPO is denoted as follows,

\begin{dmath}
J (\theta) = 
    \mathbb{E} [ 
        \min( \\
            r(\theta) \ \hat{A}_{\theta_{old}}(s, a),\\
            {clip}(r(\theta), 1 - \epsilon, 1 + \epsilon)     \hat{A}_{\theta_{old}}(s, a))
    ]
\end{dmath}

\noindent where 
\begin{itemize}
  \item $r(\theta)$ is the probability ratio between old and new policies ${\pi_\theta(a \vert s)} / {\pi_{\theta_{old}}(a \vert s)}$.
  \item $\pi_\theta(a \vert s)$ is a policy, given parameter $\theta$, state $s$ and action $a$. 
  \item  ${clip}(r(\theta), 1 - \epsilon, 1 + \epsilon)$ clips $r(\theta)$ to be in the range of $1+\epsilon$ and $1-\epsilon$.
  \item  $\hat{A}(s, a)$ is an estimate of the advantage function $A(s, a) = Q(s, a) - V(s)$, given action-value function $Q(s, a)$ and state-value function $V(s)$.
\end{itemize}

Typically $J (\theta)$ is updated via stochastic gradient ascent with  an optimiser such as Adam\cite{Kingma2014Adam:Optimization}.

\subsection{TrueSkill\texttrademark \ Ranking System}
To measure the competitiveness of the learned RL agents, the TrueSkill\texttrademark ~ranking system \cite{Herbrich2007TrueSkill:System} was used. The TrueSkill\texttrademark ~ranking system is a skill based ranking system that quantifies a players’ rating using the Bayesian inference algorithm. This system has been frequently used in many different multiplayer games and sports applications \cite{Tarlow2014KnowingUncertainty}. Although It also works well with $N$-player team games and free-for-all games, we focus our attention on the simplest case, a two-player match.

Each rating is characterised by a Gaussian distribution with mean $\mu$ and standard deviation $\sigma$. These values are updated based on the outcome of a game with the following update equations,

\begin{flalign} 
&\mu_{winner} \leftarrow \mu_{winner} + \frac{\sigma^{2}_{winner}}{c} \cdot v (\frac{\mu_{winner} - \mu_{loser}}{c},\frac{\epsilon}{c}) \\
&\mu_{loser} \leftarrow \mu_{loser} + \frac{\sigma^{2}_{loser}}{c} \cdot v (\frac{\mu_{winner} - \mu_{loser}}{c},\frac{\epsilon}{c}) \\
&\sigma_{winner} \leftarrow \sigma_{winner} \cdot [ 1 - \frac{\sigma_{winner}}{c^2} \cdot w (\frac{\mu_{winner} - \mu_{loser}}{c},\frac{\epsilon}{c}) ] \\
&\sigma_{loser} \leftarrow \sigma_{loser} \cdot [ 1 - \frac{\sigma_{loser}}{c^2} \cdot w (\frac{\mu_{winner} - \mu_{loser}}{c},\frac{\epsilon}{c}) ] \\
&c^2 = 2\beta^2 + \sigma^2_{winner} + \sigma^2_{loser} 
\end{flalign}

\noindent where $\epsilon$ is a configurable parameter that should be adjusted accordingly to the likeliness of a draw, and $\beta$ is the variance of the performance around the skill of each player. $v$ and $w$ are functions that are designed so that weighting factors are roughly proportional to the uncertainty of the winner/loser vs. the total sum of uncertainties. We refer the reader to the original paper \cite{Herbrich2007TrueSkill:System}  for further explanation. Finally, a so-called conservative skill estimate can be calculated by $\mu - k * \sigma$, where $k$ is usually set to 3. 

\subsection{Social Network Analysis}\label{social network analysis}
To analyse the intelligence of coordinated RL agents and compare their characteristics with real-world data, an analysis framework that is not influenced by physical differences between simulations and the real-world is necessary. Passes do not rely on individual physical ability and is an important component of teamplay. Therefore we focus on social network analysis (SNA) of passes. 

A pass network is a weighted directed graph that considers the direction and frequency of passes between two players. It takes the form of an adjacency matrix $A$ and weight matrix $W$. $A_{ij}$ represents the number of passes from player $i$ to player $j$, and $W_{ij}$ is simply $1/A_{ij}$ if $i\neq j$ or $0$ otherwise. Below, we explain the three metrics used in this paper.

\noindent\textbf{Closeness Centrality.} Closeness is calculated by computing the sum of all the geodesic (shortest) paths between the node $v$ and all other nodes $w \in V$ in the following equation. 
% This score indicates how easy it is for a player to be connected with teammates. Therefore a high closeness score indicates that a player is well-connected within the team. $\sigma_{vw}$ is defined as the shortest distance between nodes $v$ and $w$.

\begin{dmath}
Closeness(v) = \frac{1}{\sum_{w \in V}\sigma_{vw}}
\end{dmath}
\noindent where $\sigma_{vw}$ is defined as the shortest distance between nodes $v$ and $w$.
This score indicates how easy it is for a player to be connected with teammates. Therefore a high closeness score indicates that a player is well-connected within the team. 

\noindent\textbf{Betweenness Centrality.} Betweenness is calculated by counting the total numbers of geodesic paths linking $v$ and $w$ and the number of those paths that intersect a node $n$ in the following equation.  

\begin{dmath}
Betweeness(v) = \sum_{s \neq v \in V}\sum_{t \neq v \in V} \frac{\sigma_{st}(v)}{\sigma_{st}}
\end{dmath} 

\noindent where $\sigma_{st}(v)$ is the number of shortest paths from node $s$ to node $t$ that passes node $v$.
This score indicates how players acts as a bridge between passing plays, high deviation within a team may indicate well-balanced passing strategy and less dependence on a single player.

\noindent\textbf{Pagerank Centrality.} Pagerank is calculated based on the total number of passes a player made in the following equation. % and two heuristic parameters, $p$ which represents the probability a player will decide not pass the ball and $q$ which can be thought of "free popularity". These parameters are set to $p=0.85$ and $q=1$ following \cite{AStrategies}. A high pagerank score implies that the player is a popular choice for other players to pass too.

\begin{dmath}
Pagerank(v) = p \sum_{v\neq w}\frac{A_{vw}}{L_{w}^{out}}Pagerank(w)+q
\end{dmath}% \[c_{i}^{w} = \frac{1}{u_i(u_i - 1)} \sum_{j,k}\frac{\sqrt[3]{A_{ij}A_{kj}A_{ki}}}{\max(A)}\]
\noindent where $p$ represents the probability a player will decide not pass the ball and $q$ can be thought of "free popularity", both of which are heuristic parameters. These parameters are set to $p=0.85$ and $q=1$ following \cite{Pena2012AStrategies}. A high pagerank score implies that the player is a popular choice for other players to pass too.

\section{\uppercase{Proposed Analysis Framework}}

In this section, we present the details of our proposed analysis framework, which is outlined in Fig.~\ref{overview}, and the details regarding the setup of the subsequent experiments.
Our framework consists of five parts. In the first part (i), we train agents using proximal policy optimisation in the Google Research Football simulation environment. (ii) Then, we rank the agents by the TrueSkill ranking system. In the third part (iii), we extract event data concerning on-the-ball actions from the simulations and convert it into a tabular format. This format is similar to the Soccer Player Action Description Language (SPADL) but simplified to only include passes and shots. We also convert real-world football data into the same format as well. Finally, we perform (iv) correlation analysis and (v) social network analysis on the obtained data. 

\subsection{Agent Training and Ranking}\label{Agent Training and Ranking}
In order to train agents, we closely follow the setup of the baseline agents for the Google Research Football environment presented in \cite{Kurach2019GoogleEnvironment}. An agent will control a single active player at all timesteps and has the ability to switch to control any other player on the same team (excluding the goal keeper). Non-active players are controlled via another in-game rule based system. In this system, the behavior of the non-active players corresponds to simple actions such as running towards the ball when not in possession, or move forward together with the active player when in possession. Hence, the players can be regarded as being centrally controlled. In this paper we consider multi-agent RL to be out of scope and hope to pursue such a setup in the future. 

\subsubsection{Deep RL Implementation}
\begin{figure}[t]
\includegraphics[width=\linewidth]{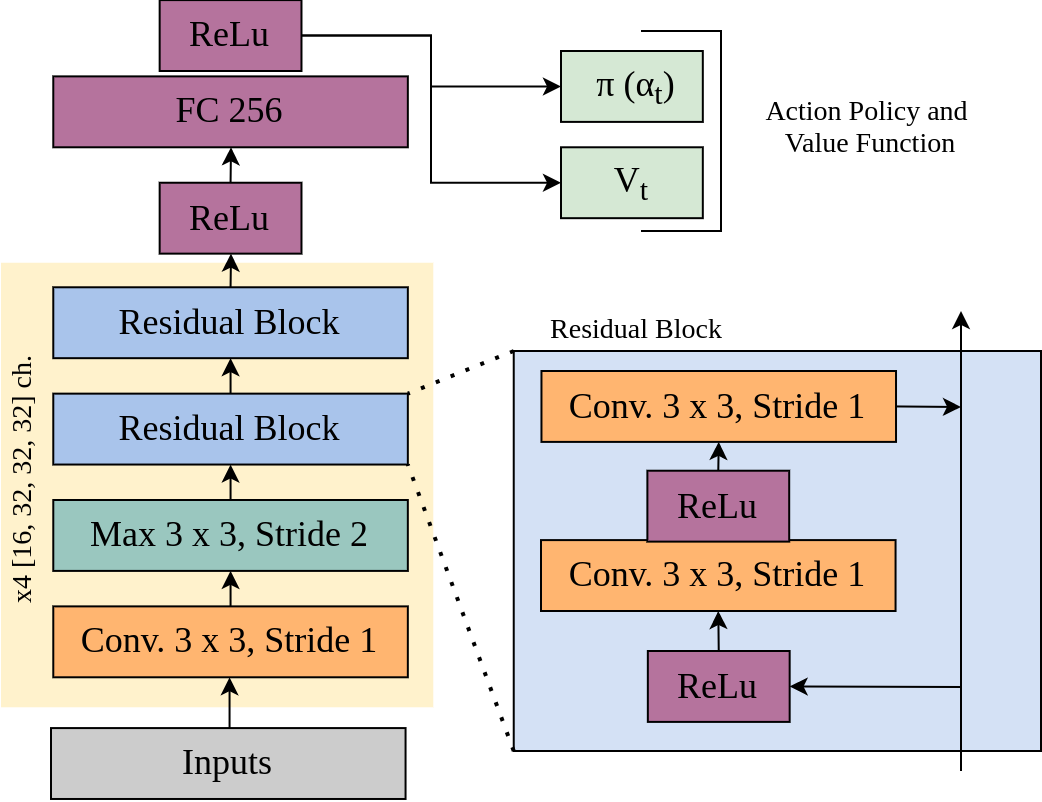}
\caption{An overview of the architecture used for the PPO agents \cite{Kurach2019GoogleEnvironment}. A stack of four previous frames (see Fig.~\ref{smm}) is used as input.} \label{architecture}
\end{figure}

The training pipeline is as follows. First, we reproduce the results presented in \cite{Kurach2019GoogleEnvironment} by using the same hyper-parameter/training setup. The Deep RL agent uses the PPO algorithm \cite{Schulman2017ProximalAlgorithms} as described in \S3.1, with an Impala policy \cite{Espeholt2018IMPALA:Architectures}. The architecture is available Fig.~\ref{architecture}.

\begin{figure}[t]
\includegraphics[width=\linewidth]{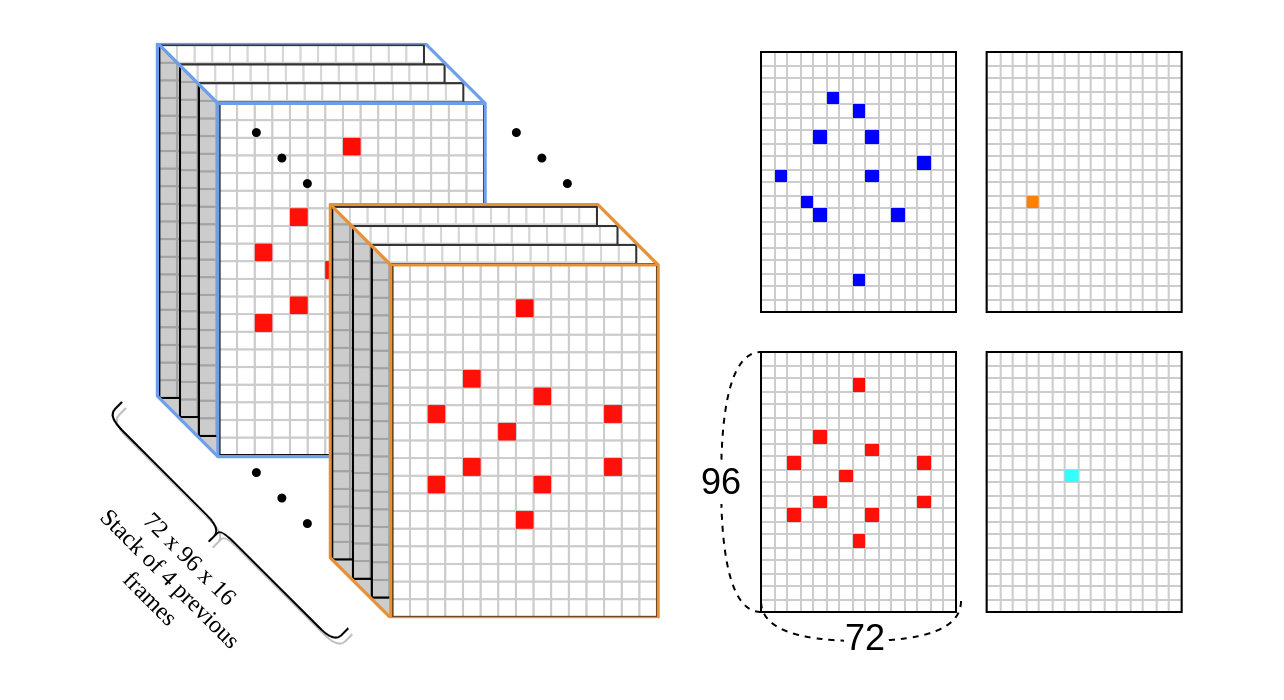}
\caption{Overview of super mini map
\cite{Kurach2019GoogleEnvironment}. Left: A stack of four previous frames used as input for the CNN. Right: A visualisation of an example stacked mini map representation.} \label{smm}
\end{figure}

Each state of the simulation is represented by a Super Mini Map (SMM) based on \cite{Kurach2019GoogleEnvironment}. The SMM consists of four $72 \times 96$ matrices, each a binary representation of the locations of the home team players, the away team players, the ball and the active player, respectively. A visualisation can be found in Fig.~\ref{smm}. The actions available\footnote{See https://git.io/Jn7Oh for a complete overview of observations and actions} to the central control agent are displayed in Table~\ref{actions}. Each movement action is sticky, therefore once executed, the action will persist until there is an explicit stop action.
% \usepackage{booktabs}
\begin{table}[!h]
\caption{Set of Actions}
\label{actions}
\begin{tabular}{ccc}
\hline
Top          & Bottom       & Left        \\
Right        & Top-Left     & Top-Right   \\
Bottom-Left  & Bottom-Right & Shot        \\
Short Pass   & High Pass    & Long Pass   \\
Idle         & Sliding      & Dribble     \\
Stop-Dribble & Sprint       & Stop-Moving \\
Stop-Sprint  & -            & -           \\
\hline
\end{tabular}
\end{table}

Rewards are based on whether a goal is conceded, scored, or neither. In addition to this goal-based reward a small "checkpoint" reward is used to aid the initial development where goals are sparse. We refer the reader to ~\cite{Kurach2019GoogleEnvironment} for a more in-depth description of possible training setups.

Based on the above setup, in this paper, we started by training for 50 million time-steps against the built-in easy, medium and hard level bots. During this phase, we noticed that the performance of the agents had not converged. Therefore, we trained an extra 50-million time-steps against the easy and medium bots and an extra 150-million time-steps against the hard-level bot. The average goal difference for the resulting agents at 50, 100 and 200 million time-steps is presented in Table~\ref{train-results}. %The average-goal difference over time-steps (the training curve) is plotted in ~\ref{rewards-over-time}. 
\begin{table}[h]
    \centering
    \caption{Average Goal Difference.}\label{train-results}
    \begin{tabular}{llll}
    \toprule
    \textbf{Bot Level} & \textbf{50M} & \textbf{100M} & \textbf{200M} \\ \midrule
    Easy      & 5.66               & 8.20                & -                   \\
    Medium    & 0.93               & 2.35                & -                   \\
    Hard      & -0.08              & 1.25                & 2.81                \\ \bottomrule
    \end{tabular}
\end{table}

\subsubsection{TrueSkill Ranking Implementation} \label{TrueSkill Ranking Implementation}
To implement the TrueSkill ranking, we create a round-robin tournament composed of 15 agents (5 from each setup, easy, medium and hard) using intermediate checkpoints saved at 20\%, 40\%, 60\%, 80\% and 100\% of training. In a single round-robin tournament, each agent plays every other agent once. We conducted a total of 50 round-robin tournaments, resulting in a total of 5250 matches. Next, we use the resulting scores of all 5250 matches to calculate a TrueSkill rating for each agent. We show the top-3 / bottom-3 ranked agents of the resulting leader-board in Table~\ref{leaderboard}. Notice the agents trained against the easy level built-in bot ranks top 1, 2 and 3. This result seems counter intuitive, since agents trained longer against stronger built-in bots should be more competitive. Therefore this suggests that there could be better training strategies. However, exploring alternative training strategies is out of scope for this work and shall be left for future work. 

\begin{table}[h]
    \centering
    \caption{TrueSkill ratings top/bottom-3}\label{leaderboard}
    \begin{tabular}{llll}
    \toprule
    \textbf{Ranking} & \textbf{Bot Level} & \textbf{Checkpoint \%} & \textbf{rating}   \\ \midrule
    1          & Easy           & 80\%                & 34.1        \\
    2          & Easy           & 100\%               & 31.5        \\
    3          & Easy           & 40\%                & 31.5        \\
    &&...&\\
    13         & Easy           & 20\%                & 8.3         \\
    14         & Hard           & 20\%                & 7.9         \\
    15         & Medium         & 20\%                & 7.0         \\
    \bottomrule
    \end{tabular}
\end{table}
\subsection{Data Extraction}\label{Data Extraction}

Action data and observation data are extracted from the games saved when calculating TrueSkill ranking. From this data, we extract all pass and shot actions and programmatically label their results based on the following events. For real-world football data, we use event-stream data for three matches from the 2019-2020 J1-League. The J1-League is the top division of the Japan professional football league.  The data was purchased from DataStadium Inc. We show the match results in Table ~\ref{game-info}. The three teams, Kashima Antlers, Tokyo FC and Yokohama F Marinos were chosen since they were the top-3 teams on the leaderboard at the time.

% \usepackage{booktabs}
\begin{table}[h]
    \caption{Details of the real-world football data used.}\label{game-info}
\begin{tabular}{@{}llll@{}}
\toprule
\textbf{Date} &\textbf{Home Team} & \textbf{Score} & \textbf{Away Team} \\ \midrule
2019/04/14    & FC Tokyo                                                     & (1-3)          &  \begin{tabular}[c]{@{}l@{}}Kashima \\ Antlers\end{tabular} \\ \midrule
2019/04/28    & \begin{tabular}[c]{@{}l@{}}Yokohama \\ F Marinos\end{tabular} & (2-1)          & \begin{tabular}[c]{@{}l@{}}Kashima \\ Antlers\end{tabular}   \\ \midrule
2019/06/29    & FC Tokyo                                                     & (4-2)          & \begin{tabular}[c]{@{}l@{}}Yokohama \\ F Marinos\end{tabular} \\ \midrule
2019/08/10    & \begin{tabular}[c]{@{}l@{}}Kashima \\ Antlers\end{tabular}   & (2-1)          & \begin{tabular}[c]{@{}l@{}}Yokohama \\ F Marinos\end{tabular} \\ \midrule
2019/09/14    & \begin{tabular}[c]{@{}l@{}}Kashima \\ Antlers\end{tabular}   & (2-0)          & FC Tokyo                                                     \\ \bottomrule
\end{tabular}
\end{table}
We also extract all pass and shot actions from this data. The results format of both simulation and real-world data is tabular and a simplified version of SPADL \cite{Decroos2019ActionsSoccer}. An explanation of the variables used in analysis is listed in Table~\ref{variables}.

\begin{table}[!h]
    \centering
    \caption{Explanation of variables used in analysis.}\label{variables}
    \begin{tabular}{l|l}
    \toprule
    \textbf{Variables} & \textbf{Explanation}  \\ \midrule
    Shots       & Number of shot attempts.        \\
    Passes      & Number of pass attempts.        \\
    PageRank    & See \S\ref{social network analysis} PageRank Centrality.        \\
    Closeness   & See \S\ref{social network analysis} Closeness Centrality.        \\
    Betweenness & See \S\ref{social network analysis} Betweenness Centrality. \\
    \bottomrule
    \end{tabular}
\end{table}

\subsection{Data Analysis}\label{Data Analysis}
Two types of football analysis are applied to the extracted data. We first focus on the finding statistics and metrics that correlate with the agent's TrueSkill ranking. For this we calculate simple descriptive statistics, such as number of passes/shots, and social network analysis (SNA) metrics, such as closeness, betweenness and pagerank. As explained in \S\ref{social network analysis}, SNA was chosen because it describes the a team ball passing strategy. Therefore it is sufficient for the analysis of central control based RL agents. We calculate Pearson correlation coefficient and $p$-value for testing non-correlation. The following criteria were used to interpret the magnitude of correlation: values less than 0.3 were interpreted as trivial; between 0.3 and 0.5 as moderate; between 0.5 and 0.7 as strong; between 0.7 and 0.9 as very strong; more than 0.9 as nearly perfect. A $p$-value less than 0.05 is considered as statistically significant, any result above this threshold will be deemed unclear. 

Our second focus is the comparison of SNA metrics between RL agents and real-world football data. By using SNA metrics, we can compare the ball passing strategy between RL agents and real-world football data. To assure a fairness, we bootstrap $N=500$ samples of passes from each team before generating a pass network to analyse. We repeat this process 50 times. Then, we conduct normality tests to determine that the distribution is Gaussian. Finally, we plot and visually inspect the distribution.

\section{\uppercase{Results and Discussion}}

In this section, we show the results of the two types of data analysis detailed in \S\ref{Data Analysis}. The first is a correlation analysis between descriptive statistics / SNA metrics and TrueSkill rankings. The second is a comparative analysis which uses SNA metrics generated from RL agents (Google Research Football) and real-world football players (2019-2020 season J1-League).

\subsection{Correlation Analysis}

For each team an agent controls, descriptive statistics and SNA metrics were calculated using the variables listed in Table~\ref{variables}. The Pearson correlation coefficients are shown in Table~\ref{sna_correlation}. 

\begin{table}[h]
\centering
\caption{Correlation coefficients and p-values for each metric. Metrics with very strong and nearly perfect correlation are emphasised in bold.}
\label{sna_correlation}
\begin{tabular}{l|l|l}
\toprule
                    \textbf{Metric} & 
                    \begin{tabular}[x]{@{}c@{}}\textbf{Correlation}\\\textbf{Coefficient}\end{tabular} 
                    & \textbf{$p$-value} \\ \midrule
Total Passes        & -0.5                    & 0.061   \\ \midrule
\textbf{Total Shots}         & \textbf{0.77}                    & \textbf{0.001}   \\ \midrule
Successful Pass Pct & 0.62                    & 0.014   \\ \midrule
Successful Shot Pct & 0.68                    & 0.005   \\ \midrule
PageRank (std)      & 0.58                    & 0.022   \\ \midrule
PageRank (mean)     & -0.05                   & 0.848   \\ \midrule
PageRank (max)      & 0.48                    & 0.068   \\ \midrule
\textbf{PageRank (min)}       & \textbf{-0.91}                   & \textbf{0.001}   \\ \midrule
Closeness (std)     & -0.54                   & 0.036   \\ \midrule
Closeness (mean)    & -0.64                   & 0.010   \\ \midrule
Closeness (max)     & -0.61                   & 0.015   \\ \midrule
Closeness (min)     & -0.66                   & 0.007   \\ \midrule
Betweenness (std)   & 0.65                    & 0.009   \\ \midrule
\textbf{Betweenness (mean)}  & \textbf{0.72}                    & \textbf{0.002}   \\ \midrule
Betweenness (max)   & 0.65                    & 0.009   \\ \midrule
Betweenness (min)   & 0.0                     & 0.0     \\ 
\bottomrule
\end{tabular}
\end{table}

\noindent As can be seen in Table~\ref{sna_correlation}, many of the descriptive statistics and SNA metrics have a strong correlation with TrueSkill rankings. We observe that "Total Shots" and "Betweenness (mean)" have a very strong positive correlation with TrueSkill rankings. On the other hand, "PageRank (min)" has a nearly perfect negative correlation.

\begin{figure}[!h]
  \centering
  \begin{minipage}[b]{0.45\textwidth}
    \includegraphics[width=\textwidth]{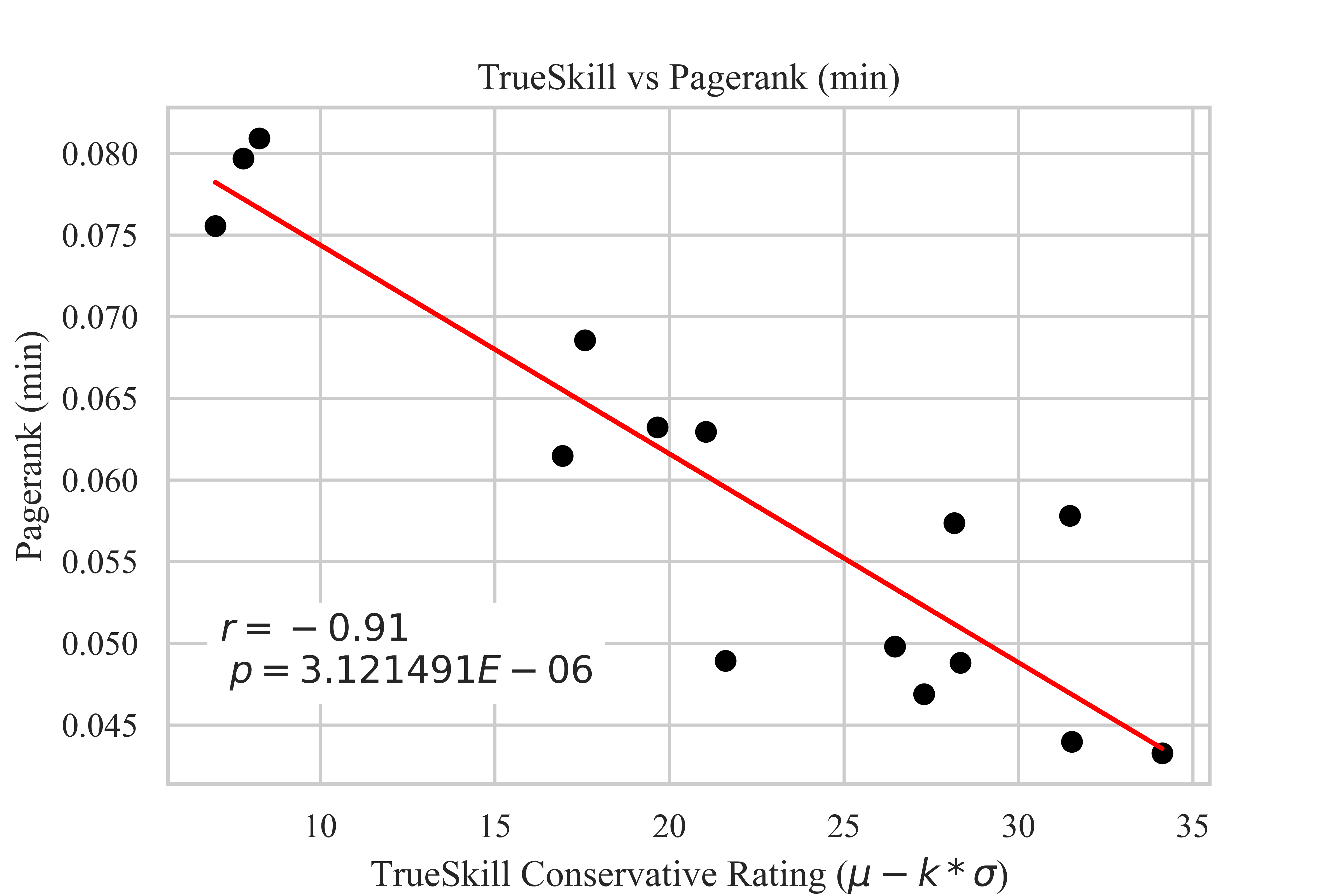}
    \caption{Pagerank aggregated by the minimum value in the network.}\label{pagerank}
  \end{minipage}
  \hfill
\end{figure}

The metric with the largest overall correlation is the pagerank aggregated by the minimum value in the network ($r=-0.91$, $p=0.001$). We present a scatter plot of this metric in Fig.~\ref{pagerank}.                 

\begin{figure}[!h]
  \centering
  \begin{minipage}[b]{0.45\textwidth}
    \includegraphics[width=\textwidth]{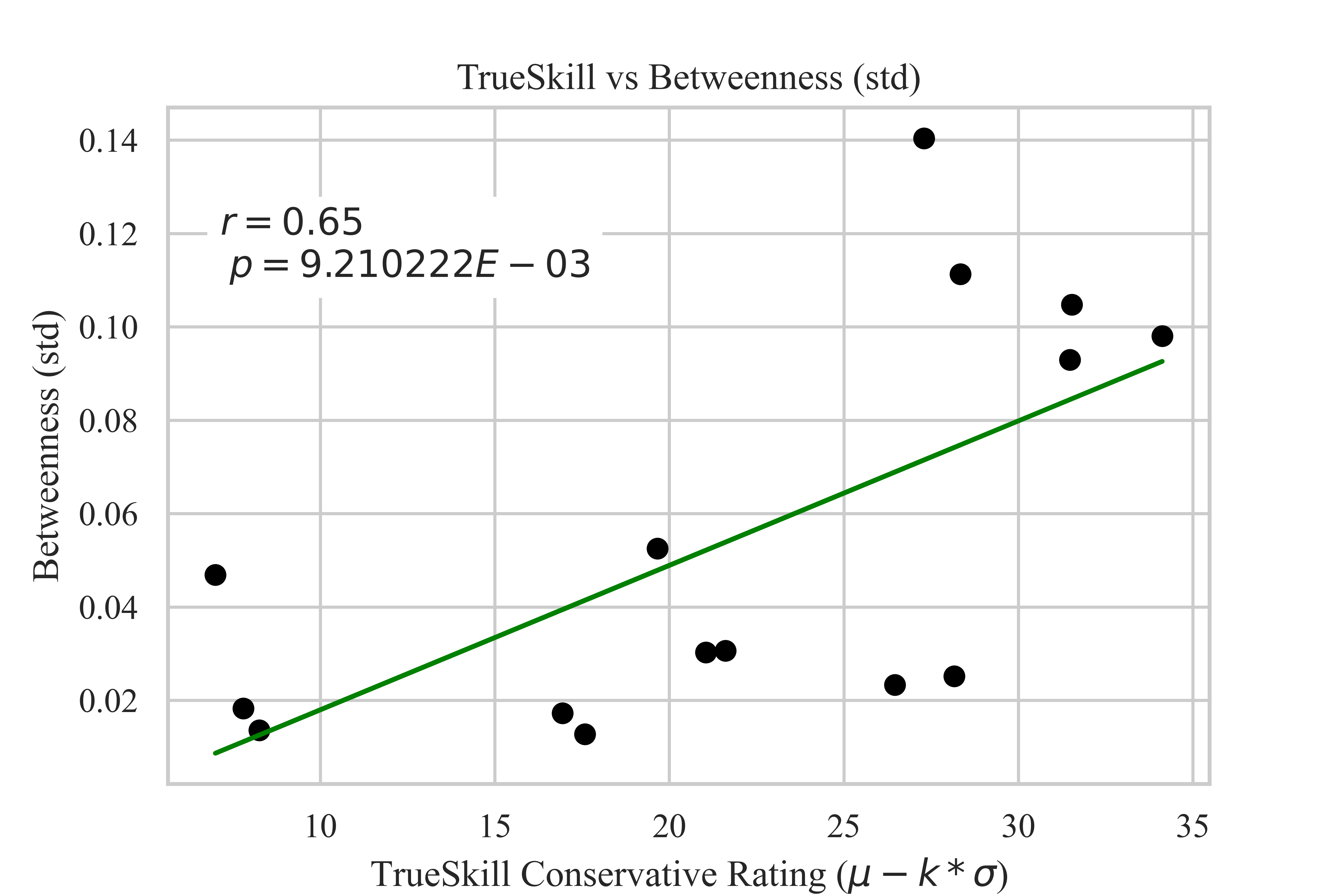}
    \caption{Betweenness aggregated by the standard deviation.}\label{betweenness}
  \end{minipage}
    \hfill
\end{figure}

Since pagerank roughly assigns to each player the probability that they will have the ball  after a arbitrary number of passes, the node with the minimum pagerank centrality is likely to be the goalkeeper, whom we assume that the agent is quickly learning to keep the ball away from. Another interesting finding is the strong positive correlation with the standard deviation of betweenness ($r=0.65$, $p=0.009$). This metric is also presented as a scatter plot in Fig.~\ref{betweenness}. 

A large variance in betweenness has been demostrated to be related with a well-balanced passing strategy and less specific player dependence \cite{Clemente2016SocialAnalysis}. It is fascinating that the agents learn to prefer a well-balanced passing strategy as TrueSkill increases. In general, most of the metrics presented in Table~\ref{sna_correlation} have either a negative or positive moderate strong correlation with $p < 0.05$.  

% \begin{figure*}[t]
% \includegraphics[width=\linewidth]{figures/sna_correlation.png}
% \caption{Correlation coefficients for various metrics.} \label{sna_correlation}
% \end{figure*}

\subsection{Comparative Analysis Between Simulated and Real-world Football}
\begin{figure}[!h]
\includegraphics[width=\linewidth]{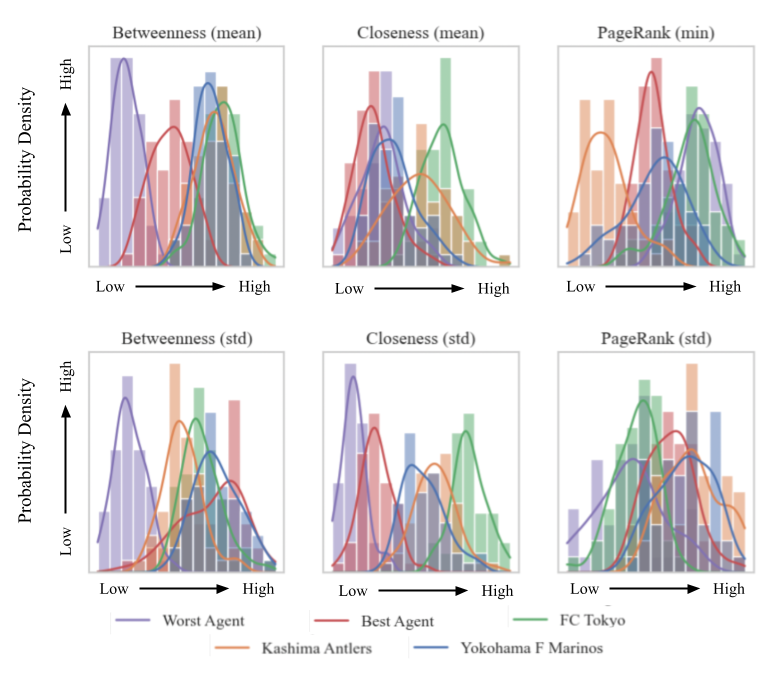}
\caption{Comparison of SNA metrics between best/worst agents and real-world football teams. } \label{comparison_dist}
\end{figure}

As exaplained in \S\ref{Data Extraction}, for each of the five real world football matches played by three teams, we calculated the distribution of SNA metrics. Distributions were calculated by bootstrapping $N=500$ samples of passes 50 times. The same procedure was taken for the matches played by the best and worst ranked agents  (see Table ~\ref{Agent Training and Ranking}). In Fig.~\ref{comparison_dist} we visualise each of the three SNA metrics aggregated by two different methods. Aggregation methods that showed strong correlations in Table \ref{sna_correlation} were chosen. The total number of passes and shots per match can not be fairly compared between RL-agents and real-world footballers because of different match lengths. 
In summary, a total of six variables were compared over five agents/teams (worst RL agent, best RL agent, FC Tokyo, Kashima Antlers and Yokohama F Marinos).

Observing this visualisation we can see that the distribution of "Betweenness (mean)", "Betweenness (std)" and "Closeness (std)" metrics for the worst agent is distant from the others. The fact that the best agent distribution of the same metric is much closer to that of J League teams implies that agent has learnt to play in a similar style through RL. However the same cannot be said for the other metrics, "Closeness (mean)", "PageRank (std)" and "PageRank (min)".

From the perspective of football analysis, the distributions of "Betweenness (std)" is very interesting. Since a high deviation in betweenness may indicate well-balanced passing strategy and less dependence on a single player, we can hypothesise that agents are learning to play a more well-balanced passing strategy similar to real-world footballers. 

Although it is difficult to interpret the results from the PageRank and Closeness metrics, it is surprising that even the worst RL agents have overlapping distributions with the real-world footballers. Considering the fact that even the worst RL agent was trained thousands of timesteps, this may be because strategies related PageRank and Closeness are easier to learn.

\section{\uppercase{Conclusions and Future work}}
In this paper, we compared the characteristics and play styles of RL agents of increasing competitiveness. As a result, we found many metrics that strongly correlate with the competitiveness (TrueSkill rating) of an agent. Another contribution in this paper, is the comparison between RL agents and real football players. Our findings suggest that an RL agent can learn to play football in similar style to that of real player without being explicitly programmed to do so.

There are many directions we can extend the research presented in this paper. In particular, we plan to work on increasing the degree of freedom within the simulations to create a more realistic environment. This can be achieved by conducting multi-agent simulation where an RL agent controls a single active player in contrast to a whole team. Another approach would be to use a less restrictive environment such as the “Humanoid Football” environment to introduce biomechanical movements. Although both approaches appear interesting, improvements in training methodology, such as imitation learning and auto-curricular learning may be required to produce adequate agents. 

We also noticed that it was difficult to use state of the art football analysis methods due to different representations of the underlying data. Since efficient representations such as SPADL already exist, we hope other researchers can build on top of these so that the community can easily take advantage of existing methods.

% (1) We compared the characteristics of football playing RL agents and real football players for the first time, thus verifying simulations as a practical approach for football analysis;
%(2) We analyzed the characteristics of football playing RL agents using football-related aggregated statistics and network analysis, which contribute to the understanding RL agents from a more complex football perspective

% \section*{ACKNOWLEDGEMENTS}
% We thanks the reviewers for their constructive suggestions and comments. We also thanks Keisuke Fuji, Kazushi Tsutsui  and Aranha Claus for helpful discussions.

\bibliographystyle{apalike}
{\small\bibliography{references}}

% \section*{\uppercase{Appendix}}

% If any, the appendix should appear directly after the
% references without numbering, and not on a new page. To do so please use the following command:
% \textit{$\backslash$section*\{APPENDIX\}}

\end{document}